# SENTENCE-LEVEL DIALECTS IDENTIFICATION IN THE GREATER CHINA REGION

Fan Xu, Mingwen Wang and Maoxi Li

School of Computer Information Engineering, Jiangxi Normal University
Nanchang 330022, China

## ABSTRACT

*Identifying the different varieties of the same language is more challenging than unrelated languages identification. In this paper, we propose an approach to discriminate language varieties or dialects of Mandarin Chinese for the Mainland China, Hong Kong, Taiwan, Macao, Malaysia and Singapore, a.k.a., the Greater China Region (GCR). When applied to the dialects identification of the GCR, we find that commonly used character-level or word-level uni-gram feature is not very efficient since there exist several specific problems such as the ambiguity and context-dependent characteristic of words in the dialects of the GCR. To overcome these challenges, we use not only the general features like character-level n-gram, but also many new word-level features, including PMI-based and word alignment-based features. A series of evaluation results on both the news and open-domain dataset from Wikipedia show the effectiveness of the proposed approach.*

## 1. INTRODUCTION

Automatic language identification of an input text is an important task in Natural Language Processing (NLP), especially when processing speech or social media messages. Besides, it constitutes the first stage of many NLP pipelines. Before applying tools trained on specific languages, one must determine the language of the text. It has attracted considerable attention in recent years [1, 2, 3, 4, 5, 6, 7, 8]. Most of the existing approaches take words as features, and then adopt effective supervised classification algorithms to solve the problem.

Generally speaking, language identification between different languages is a task that can be solved at a high accuracy. For example, Simoes et al. [9] achieved 97% accuracy for discriminating among 25 unrelated languages. However, it is generally difficult to distinguish between related languages or variations of a specific language (see [9] and [10] for example). To deal with this problem, Huang and Lee [3] proposed a contrastive approach based on document-level top-bag-of-word similarity to reflect distances among the three varieties of Mandarin in China, Taiwan and Singapore, which is a kind of word-level uni-gram feature. The word uni-gram feature is sufficient for document-level identification of language variants.

More recent studies focus on sentence-level languages identification, such as the Discriminating between Similar Languages (DSL) shared task 2014 and 2015 [7, 8]. The best system of these shard tasks shows that the uni-gram is an effective feature. For the sentence-level language identification, you are given a single sentence, and you need to identify the language.

Chinese is spoken in different regions, with noticeable differences between regions. The first difference is the character set used. For example, Mainland China and Singapore adopt simplified character form, while Taiwan and Hong Kong use complex/traditional character form, as shown in the following two examples.

    (1)Simplified character form examples
    餐厅/can ting/restaurant; 机构/ji gou/organization; 回顾/hui gu/review.
    (2)Complex/traditional character form examples
    餐廳/can ting/restaurant; 機構/ji gou/organization; 回顧/hui gu/review.

We observe furthermore that the same meaning can be expressed using different linguistic expressions in the Mainland China, Hong Kong and Taiwan variety of Mandarin Chinese. Table 1 lists some examples. As a result, the words distribution should be different in the Chinese variants spoken in the Mainland China, Hong Kong, Taiwan, Macao, Malaysia and Singapore variety or dialect[1] of Mandarin Chinese, a.k.a., the Greater China Region (GCR). Therefore, we can extract the different fine-grained representative words (n-gram with $n \leq 3$; lines 2-5 in Table 1) for the GCR respectively using Pointwise Mutual Information (PMI) in order to reflect the correlation between the words and their ascribed language varieties. Compared with English, no space exists between words in Chinese sentence. Due to the Chinese word segmentation issue, some representative words for the GCR cannot be extracted using PMI (lines 6-8 in Table 1). To expand these representative words for each dialect, we extract more coarse-grained (n-gram with $n \geq 4$) words using a word alignment technology, and then propose word alignment-based feature (dictionary for each dialect with n-gram under $n \geq 4$). In fact, the word alignment-based dictionary can extract both fine-grained representative words and coarse-grained words simultaneously.

Table 1. Some GCR word alignment word set examples.

| Chinese Mainland | Hong Kong | Taiwan |
|---|---|---|
| 出租车<br>(chu zu che/taxi) | 的士<br>(di shi/taxi) | 计程车<br>(ji cheng che/ taxi) |
| 查找<br>(cha zhao/find) | 寻找<br>(xun zhao/find) | 寻找<br>(xun zhao/find) |
| 生态圈<br>(sheng tai quan/ecosystem) | 生态系<br>(sheng tai xi/ecosystem) | 生态系<br>(sheng tai xi/ecosystem) |
| 方便面<br>(fang bian mian/instant noodles) | 即食面<br>(ji shi mian/instant noodles) | 速食面<br>(su shi mian/instant noodles) |
| 乒乓球拍<br>(ping pang qiu pai/table tenis bat) | 乒乓球拍<br>(ping pang qiu pai/table tenis bat) | 桌球拍<br>(zhuo qiu pai/table tenis bat) |
| 人机界面<br>(ren ji jie mian/human interface) | 人机介面<br>(ren ji jie mian/human interface) | 人机介面<br>(ren ji jie mian/human interface) |
| 五角大楼<br>(wu jiao da lou/pentagon) | 五角大厦<br>(wu jiao da sha/pentagon) | 五角大厦<br>(wu jiao da sha/pentagon) |

---

[1] For the sake of simplicity, we refer to both languages and language varieties as languages.

The above observation indicates that character form, PMI-based and word alignment-based information are useful information to discriminate dialects in the GCR. In order to investigate the detailed characteristics of different dialects of Mandarin Chinese, we extend 3 dialects in Huang and Lee [3] to 6 dialects. In fact, the more dialects there are, the more difficult the dialects discrimination becomes. It also has been verified through our experiments. Very often, texts written in a character set are converted to another character set, in particular on the Web. This makes the character form feature unusable. In order to detect dialects for those texts, we convert texts in traditional characters to simplified characters in order to investigate the effectiveness of linguistic and textual features alone. Due to these characteristic of Chinese, current methods do not work for the specific GCR dialects.

Evaluation results on our two different 15,000 sentence-level news dataset and 18,000 sentence-level open-domain dataset from Wikipedia show that bi-gram, character form, PMI-based and word alignment-based features significantly outperform the traditional baseline systems using character and word uni-grams.

The main contributions of this paper are as follows:

(1) We find character-level bi-gram and word segmentation based features work better than traditional character-level uni-gram feature in the dialects discrimination for the GCR;

(2) Some features such as character form, PMI-based and word alignment-based features can improve the dialects identification performance for the GCR.

The remainder of the paper is organized as follows. Section 2 presents the state-of-the-art approaches in the language identification field. Section 3 describes the main features used in this paper. The dataset collection and experiment results are shown in Section 4, and we conclude our paper in Section 5.

## 2. RELATED WORK

A number of studies on identification of similar languages and language varieties have been carride out. For example, Murthy and Kumar [1] focused on Indian languages identification. Meanwhile, Ranaivo-Malancon [2] proposed features based on frequencies of character n-grams to identify Malay and Indonesian. Huang and Lee [3] presented the top-bag-of-word similarity based contrastive approach to reflect distances among the three varieties of Mandarin in Mainland China, Taiwan and Singapore. Zampieri and Gebre [4] found that word uni-grams gave very similar performance to character n-gram features in the framework of the probabilistic language model for the Brazilian and European Portuguese language discrimination. Tiedemann and Ljubesic [5]; Ljubesic and Kranjcic [6] showed that the Naïve Bayes classifier with uni-grams achieved high accuracy for the South Slavic languages identification. Grefenstette [11]; Lui and Cook [12] found that bag-of-words features outperformed the syntax or character sequences-based features for the English varieties. Besides these works, other recent studies include: Spanish varieties identification [13], Arabic varieties discrimination [14, 15, 16, 17], and Persian and Dari identification [18].

Among the above related works, study [3] is the most related work to ours. The differences between study [3] and our work are two-fold:

(1)They focus on document-level varieties of Mandarin in China, Taiwan and Singapore, while we deal with sentence-level varieties of Mandarin in China, Hong Kong, Taiwan, Macao, Malaysia and Singapore. In order to investigate the detailed characteristic of different dialects of Mandarin Chinese, we extend dialects in Huang and Lee [3] to 6 dialects. Also, the more dialects there are, the more difficult the dialects discrimination becomes.

(2) The top-bag-of-word they proposed in Huang and Lee [3] is word uni-gram feature essentially. While in this paper, besides the traditional uni-gram feature, we propose some novel features, such as character form, PMI-based and word alignment-based features.

## 3. DIALECTS CLASSIFICATION MODELS

In this section, we recast the sentence-level dialects identification in the GCR as a multiclass classification problem. Below we will describe some common features in the general language (unrelated languages or different languages) identification as well as some novel features such as character form, PMI-based and word alignment-based features. These features are fed into a classifier to determine the dialect of a sentence.

### 3.1 Character-level Features

In this section, we represent the N-gram features and character form features.

#### 3.1.1 N-gram Features

According to the related works [4, 5, 6], word uni-grams are effective features for discriminating general languages. Compared with English, no space exists between words in Chinese sentence. Therefore, we use character uni-grams, bi-grams and tri-grams as features. However, Huang and Lee [3] did not use character-level n-grams.

#### 3.1.2 Character Form Features

Due to various historical reasons, there are many different linguistic phenomena and expression variances among the GCR. As mentioned earlier, Mainland China, Malaysia and Singapore adopt the simplified character form, while Hong Kong, Taiwan and Macao use the complex/traditional character form. This kind of information is very helpful to identify sentence-level dialects in the GCR.

Motivated by the above observations, we first construct a complex/traditional Chinese dictionary with 626 characters crawled from the URL.[2] These characters have been simplified. Thus they make the strongest differences between two character sets. Then we generate the character form based feature as a Boolean variable to detect whether the Chinese sentence contain any word in the traditional dictionary. Table 2 lists some complex/traditional character examples. Generally, we can know that the complex character form occurs mostly in the strict genre i.e. news text. Thus, this kind of information is useful to discriminate dialects in the GCR.

Table 2. Some complex character.

| Chinese character | Pinyin | English |
|---|---|---|
| 邊 | bian | side |
| 罷 | ba | stop |
| 車 | che | car |
| 鬥 | men | door |

---

| 過 | guo | pass |
| 龍 | long | dragon |
| 馬 | ma | horse |
| 貓 | mao | cat |
| 憑 | ping | lean on |
| 萬 | wan | million |

## 3.2 Word-level Features

In this section, we represent the word segmentation, PMI and word alignment features.

### 3.2.1 Word Segmentation Features

Chinese word segmentation [19] is a vital pre-processing step before Chinese information processing. As we mentioned earlier, different words may be used to express the same meaning in different dialects. Therefore, words are also useful features for dialect detection. These features have been successfully used in Huang and Lee. Thus, we firstly conduct word segmentation using ICTCLAS [3] Chinese word segmentation package which can handle both simplified and traditional/complex characters for the Chinese sentence, and then extract each word uni-gram to generate word segmentation feature vector.

### 3.2.2 PMI Features

Once a sentence is segmented into words, we adopt Pointwise Mutual Information (PMI) to determine the relationship between the words and theirs ascribed language varieties. PMI for a word only used in a dialect will be high, while the one used in all the dialects will be low.

Specifically, we calculate the relationship between words and theirs dialect type by Equation (1) as follows:

$$pmi(w_i, I_j) = \log \frac{p(w_i, I_j)}{p(w_i)p(I_j)} \qquad (1)$$

where $w_i$ indicates any word in the corpus and $I_j$ a dialect. $p(w_i)$ stands for the ratio of the frequency of a word in the corpus to the total number of words, $p(I_j)$ means the ratio of the frequency of words in the documents using dialect $j$ to the total number of words in the corpus, $p(w_i, I_j)$ indicates for the account of the frequency of the word $i$ occurs in the documents using dialect $j$ and the total number of words in the corpus. Then, we can generate different word set for each dialect of the GCR, and yield PMI-based feature according to the word set. For example, if a word in a sentence occurs among in the word set of Mainland China, Taiwan and Singapore, thus the value of PMI-based feature is MC_TW_SGP (MC stands for Mainland China, TW refers to Taiwan, and SGP means Singapore). In fact, we can take the PMI-based feature as a way of weighting the word segmentation-based features.

### 3.2.3 Word Alignment Features

As mentioned earlier, for a single semantic meaning, various linguistic expressions exist in the GCR. Then, how to align different coarse-grained expressions (n-gram with n≥4; lines 6-8 in Table1) of the same meaning for each dialect is a vital problem. We can generate dictionary for each dialect as an expansion of PMI-based word set.

---

[3] http://ictclas.nlpir.org/downloads

Intuitively, according to word alignment problem in the machine translation, given a source sentence $e$ consisting of words $e_1$, $e_2$,…, $e_l$ and a target sentence $f$ consisting of words $f_1$, $f_2$,…, $f_m$, we need to infer an alignment $a$, a sequence of indices $a_1$, $a_2$,…, $a_m$ which indicates the corresponding source word $e_{ai}$ or a null word. Therefore, we can recast the coarse-grained expressions extraction in the GCR as a word alignment problem in the statistic machine translation.

Specifically, we firstly crawl about 30 million parallel sentence pairs between Mainland China and each other dialect in the GCR from parallel texts in the news and Wikipedia website. The corpus collected for the word alignment features different from the test data as described in the subsection 4.1, which are not just the same texts converted from traditional characters to simplified characters, or vice versa, and then extract the word alignment using GIZA++ [20]. After removing the Longest Common Subsequence (LCS) ([21]) of words, we can extract the different linguistic expressions mapping between Mainland China and each other dialect in the GCR. Then, we generate about 12,374 parallel word set for each dialect of the GCR, and yield word alignment-based feature according to the word set. For example, if a word occurs in the word set of Mainland China, Singapore and Malaysia, then we set the value of word alignment-based feature as MC_SGP_MAL (MC stands for Mainland China, SGP means Singapore, and MAL refers to Malaysia). To be specific, Figure 1 shows an example to extract word alignment from two parallel sentences. As shown, we can extract "人机界面/ren ji jie mian/human interface" and "人机介面/ren ji jie mian/human interface" to generate dictionary for each dialect.

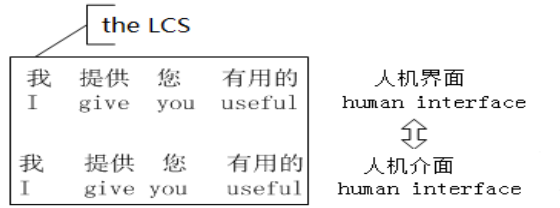

Figure 1. A parallel sentence pairs written in simplified script
for Chinese mainland and traditional script for Hong Kong.

## 3.3 Classifier

After extracting the above proposed features, in order to do fair comparison with the baseline systems (Section 4.1), we train a single or combined multiclass linear kernel support vector machine using LIBLINEAR [22] with default parameters such as verbosity level with 1, trade-off between training error and margin with 0.01, slack rescaling, zero/one loss. Also, due to large number of the parameters of SVM, we do not tune them on the development sets. According to previous studies, SVM were well-suited to high-dimensional feature spaces; SVM has shown good performance in many other language identification work [10, 23]. Therefore, we adopt SVM to discriminate sentence-level dialects for the GCR. Besides, we trained maximum entropy and naïve Bayes classifiers, but the results are much worse than SVM. We also trained some other kernel function of SVM with polynomial, radial basis function, and sigmoid, but the linear kernel gets the best results. Consequently, we only report the results with SVM using linear kernel function in the Section 4.

# 4. EXPERIMENTS

In this section, we first introduce the experimental settings, and then evaluate the performance of our proposed approach for identifying dialects in the GCR.

## 4.1 Experimental Settings

**Dataset**: We crawl our sentence-level dialect data set from news websites[4] and Wikipedia using the jsoup[5] utility. After removing the useless sentence, such as the English sentences, the English words account 50% in total and the sentences including less than 15 words, we obtained 27,679 news sentences in total (3,452 sentences for Macao, 5,437 sentences for Mainland China, 5,816 sentences for Hong Kong, 5,711 sentences for Malaysia, 4,672 sentences for Taiwan, and 2,591 sentences for Singapore) . In order to balance these sentences for each dialect, we random select 2,500 sentences for each dialect in the GCR, thus we generate 15,000 sentences for the GCR in total. Similarly, we also extract 18,000 Wikipedia sentences (6,000 sentences for each dialect, including Mainland China, Hong Kong and Taiwan in the GCR). All sentences from the same website can be automatically annotated with a specific dialect type. There are no duplicates in the dataset. For evaluation, we adopt 5-cross validation for the two datasets.

For the news dataset, we generate three scenarios:

(1) **6-way detection**: The dialects of Mainland China, Hong Kong, Taiwan, Macao, Malaysia and Singapore are all considered;
(2) **3-way detection**: We detect dialects of Mainland China, Taiwan and Singapore as in Huang and Lee [3];
(3) **2-way detection**: We try to distinguish between two groups of dialects, the ones used in Mainland China, Malaysia and Singapore using simplified characters, and the ones used in Hong Kong, Taiwan and Macao using traditional characters.

For the Wikipedia dataset, we also generate two similar scenarios:

(1) **3-way detection**: We detect dialects of Mainland China, Hong Kong and Taiwan;
(2) **2-way detection**: We try to distinguish between two groups of dialects, the ones used in Mainland China using simplified characters, and the ones used in Hong Kong and Taiwan using traditional characters.

**Baseline system 1:** As mentioned in Section 2, we take the Huang and Lee [3]'s top-bag-of-word similarity-based approach as one of our baseline system. We re-implement their method in this paper using the similar 3-way news dataset.

**Baseline system 2:** Another baseline, word uni-gram based feature for English using SVM classifier, was proposed by Purver [10], which have been verified effective in the DSL shared task 2015.

## 4.2 Experimental Results

In this section, we report the experiment results for the dialects identification for the GCR on both news and Wikipedia dataset.

---

[4] http://news.sina.com.cn for the Mainland China, http://www.takungpao.com.hk  for the Hong Kong, http://big5.taiwan.cn for the Taiwan,  http://www.cyberctm.com  for the Macao,  http://www.sinchew.com.my for the Malaysia, and http://www.zaobao.com for the Singapore.
[5] http://jsoup.org/

### 4.2.1 Results on News Dataset

Table 3 shows the experimental results for the dialect identification in the GCR.

**(1) Single features**

If we use a single type of feature, we can see that the uni-gram feature (baseline system 2) is not the best one for Chinese dialect detection in the GCR, although it has been found effective for English detection in previous studies in the DSL shared task. Instead, bi-gram and word segmentation based features are better than uni-gram one. Both of the proposed bi-gram and word segmentation based features significantly outperforms the baseline systems with p<0.01 using paired t-test for significance. Also the bi-gram and word segmentation based features are better than the Huang and Lee [3]'s method (baseline system 1) for 6-way, 3-way and 2-way dialect identification in the GCR. Obviously, the random method does not work for the GCR dialect identification.

**(2)Bi-gram vs tri-grams and uni-gram**

Besides, the performance of bi-gram outperforms tri-gram and uni-gram. We explain this by the fact that there are much sparser in tri-grams than in bi-grams. Another explanation is that most Chinese words are formed of two characters. Bi-grams can better capture the meaningful words than tri-grams. This observation is consistent with the observation in Chinese information retrieval: Nie et al. [24] found that character bi-grams work equally well to words for Chinese information retrieval. We have the same observation in Table 3 (bi-gram vs. word segmentation).

**(3)Linguistic and alignment features**

According to Table 3, we also observe that the character form features are useful for 2-way GCR dialect classification, which verifies our observation. Also, the proposed word segment-based feature is effective for the 2-way dialect identification, which yields 98.42% accuracy.

**(4)Combined features**

For the combined features, we can know that the character form, PMI and word alignment based features can improve the language identification in the GCR. They can be successfully integrated into the effective bi-gram features. As shown in Table 3, PMI-based feature can bring performance improvement by 1.6% for the 6-way dialect identification. After integrating the novel 3 features together, we get the final best performance with 90.91% for 3-way dialects identification in the GCR. The combined features significantly outperform the bi-gram with p<0.01 using paired t-test for significance, which shows the effectiveness of our novel features.

Table 3. Accuracy using different features on the news dataset. Performance that is significantly superior to baseline systems (p<0.01, using paired t-test for significance) is denoted by *.

| Features/Systems | 6-way | 3-way | 2-way |
|---|---|---|---|
| **Baseline systems** | | | |
| random | 16.67 | 33.33 | 50.00 |
| baseline system 1; Huang and Lee [3] | 66.67 | 66.67 | 83.33 |
| uni-gram (baseline system 2; Matthew Purver [10]) | 74.59 | 85.90 | **98.51** |
| **Single feature** | | | |
| bi-gram | 80.13* | 89.25* | 98.24 |
| tri-gram | 74.93 | 86.43 | 94.46 |

| word segmentation | 78.25* | 88.41* | 98.42 |
|---|---|---|---|
| character form | 19.18 | 66.36 | 94.36 |
| PMI | 28.39 | 60.01 | 79.75 |
| Word alignment | 17.97 | 42.89 | 62.19 |
| **Combined features** | | | |
| bi-gram + character form | 80.25* | 89.55* | 97.09 |
| bi-gram + PMI | 81.71* | 89.56* | 98.13 |
| bi-gram + word alignment | 80.12* | 88.07* | 97.79 |
| bi-gram + character form + PMI | **82.00*** | 90.87* | 97.50 |
| bi-gram + PMI + word alignment | 81.79* | 89.61* | 97.93 |
| bi-gram + character form + word alignment | 80.43* | 89.19* | 97.13 |
| bi-gram + character form + PMI + word alignment | **82.00*** | **90.91*** | 97.49 |

### (5)Detailed dialect identification

More specifically, the accuracy of our best feature (bi-gram + character form + PMI + word alignment) for each dialect identification in the GCR for the 6-way classification is reported in Table 4.

Table 4. Accuracy for each dialect in the GCR using feature of
bi-gram + character form + PMI features + word alignment.

| Dialect | accuracy |
|---|---|
| Macao | 89.16 |
| Singapore | 84.28 |
| Mainland China | 81.64 |
| Hong Kong | 81.64 |
| Taiwan | 81.20 |
| Malaysia | 74.08 |

As shown, we gain the best identification performance for Macao, while the accuracy of Malaysia is the worst one. We observe much noise among the texts in Malaysia. Most sentences in Malaysia has the English words account 10% in total. Thus, how to crawl large numbers of sentences with high quality for a dialect will be one of our future works.

To be more specific, we list the confusion matrix for each dialect in the GCR in Table 5. As shown, most instances have been correctly classified. Due to the challenge of discrimination for the closely related languages in the GCR, some instances still have been falsely classified. As shown, we can know that the instances falsely classified from dialect Hong Kong to Taiwan (82) is similar to those from dialect Taiwan to Hong Kong (75). Thus we extract another open-domain dataset from Wikipedia for Hong Kong, Taiwan and Mainland China. Due to the Wikipedia sentences are parallel in two languages, we take Mainland China sentence as a bridge to extract the word alignment-based features, and further evaluation is listed in subsection 4.2.2.

Table 5. Confusion matrix for each dialect in the GCR using bi-gram + character form +PMI features; the largest falsely classified dialect is reported in bold.

| Dialect | Macao | Mainland China | Hong Kong | Malaysia | Taiwan | Singapore |
|---|---|---|---|---|---|---|
| Macao | 455 | 5 | **24** | 3 | 10 | 3 |
| Mainland China | 2 | 403 | 4 | 22 | 0 | **69** |
| Hong Kong | 23 | 2 | 390 | 2 | **82** | 1 |
| Malaysia | 4 | **48** | 2 | 394 | 4 | **48** |
| Taiwan | 15 | 1 | **75** | 0 | 408 | 1 |
| Singapore | 0 | **58** | 0 | 12 | 0 | 430 |

#### 4.2.2 Results on Wikipedia Dataset

As shown in Table 3, character form based features are very effective (94.36% for 2-way dialects classification). Similar to Huang and Lee [3]'s work, in order to eliminate the trivial issue of character encoding (simplified and traditional character), we convert Taiwan and Hong Kong texts to the same simplified character set using Zhconvertor[6] utility to focus on actual linguistic and textual features.

Table 6 shows the experimental results for the dialect identification in the GCR. As shown, again, the bi-gram features work better than both uni-gram and tri-gram features on Wikipedia dataset. Also, the word alignment-based features can contribute about 3.32% performance improvement. This also confirms our intuition that the word alignment-based information is helpful to discriminate dialects in the GCR, which shows the effectiveness of both fine-grained and coarse-grained characteristic of word alignment based features.

In order to generate word alignment word sets, as mentioned in the Section 1, the sentences are parallel for Mainland China, Hong Kong and Taiwan. After converting them to the same character set, the difference among them is subtle. Therefore, both the PMI-based feature and character form feature will be invalid in this situation. PMI depend so much on the use of a particular character set shows that it is correlated with other knowledge sources and it has been well defined. It also shows that the dialect identification on parallel sentences with same character set for the GCR is a challenging task. The reason why the performance on Wikipedia is much lower than on the news dataset is listed as follows. The texts in the Wikipedia are not written in pure dialects (maybe as mostly translated from English) or topic information biased good results achieved for the news data set, i.e. topics discussed in different news make them different. The word alignment based feature improve the performance about 3.32%, Also, the 2-way classification results shows the proposed bi-gram and word alignment based features are quite promising.

Table 6. Accuracy using different features on the Wikipedia dataset. Performance that is significantly superior to baseline systems (p<0.01, using paired t-test for significance) is denoted by *.

| Features/Systems | 3-way | 2-way |
|---|---|---|
| **Baseline systems** | | |
| baseline system 1; Huang and Lee [3] | 66.67 | 66.67 |
| uni-gram (baseline system 2; Matthew Purver [10] | 49.95 | 74.90 |
| **Single features** | | |
| bi-gram | 57.02*[uni-gram] | 77.62* |
| tri-gram | 50.69 | 70.37 |
| word segmentation | 55.50*[uni-gram] | 77.47* |
| word alignment | 33.38 | 66.67 |
| **Combined features** | | |
| bi-gram + word alignment | 60.34*[uni-gram] | **77.79*** |

## 5. CONCLUSIONS AND FUTURE WORK

In this paper, we study the problem of dialect identification for Chinese. We found that the uni-gram is commonly used in the previous work, showing very good results for European languages. Unlike the European languages, words in Chinese are not separated by spaces. Therefore, a naive

---

adaptation of the uni-gram features to Chinese character uni-gram, does not work well. However, longer elements such as character bi-grams and segmented words work much better. This indicates that such longer units are more meaningful in Chinese and can better reflect the characteristics of a dialect. In addition, we also proposed new features based on PMI and word alignment. These features are also shown useful for Chinese dialect identification.

In future work, we would like to explore more features, and test other classifiers. Furthermore, we will finally investigate how dialect identification can help other NLP tasks.

## ACKNOWLEDGEMENTS

The authors would like to thank the anonymous reviewers for their comments on this paper. This research was supported by the Research Project of State Language Commission under Grant No.YB125-99, the National Natural Science Foundation of China under Grant No.61402208, No.61462045 and No.61462044, and the Natural Science Foundation and Education Department of Jiangxi Province under Grant No. 20151BAB207027 and GJJ150351.

**Authors**

**Fan Xu** holds a Doctoral Degree (Ph.D.) in Computer Science from Soochow University, China. His areas of research interest includes Natural Language Processing, Chinese Information Processing, Discourse Analysis, and Speech Recognition. At present he is working as Lector, School of Computer Information Engineering, Jiangxi Normal University, China. He is member of various professional bodies including ACL, IEEE, and ACIS.

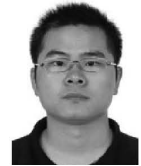

**Mingwen Wang** holds a Doctoral Degree (Ph.D.) in Computer Science from Shanghai Jiaotong University, China. His areas of research interest includes Machine Learning, Information Retrieval, Natural Language Processing, Image Processing, and Chinese Information Processing. At present he is working as Professor, School of Computer Information Engineering, Jiangxi Normal University, China. He is member of various professional bodies including ACL, IEEE, CCF, and ACIS.

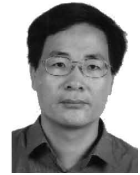

**Maoxi Li** holds a Doctoral Degree (Ph.D.) in Computer Science from Chinese Academy of Sciences. His areas of research interest includes Machine Translation and Natural Language Processing. At present he is working as Associate Professor, School of Computer Information Engineering, Jiangxi Normal University, China.

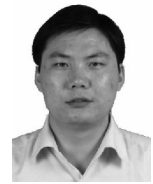